%% file: main.tex
\documentclass[letterpaper]{article} 
\usepackage{aaai2026}  
\usepackage{times}  
\usepackage{helvet}  
\usepackage{courier}  
\usepackage[hyphens]{url}  
\usepackage{graphicx} 
\usepackage{natbib}  
\usepackage{caption} 
\usepackage{algorithm}
\usepackage{algorithmic}

\usepackage{newfloat}
\usepackage{listings}
\DeclareCaptionStyle{ruled}{labelfont=normalfont,labelsep=colon,strut=off} 
\floatstyle{ruled}
\newfloat{listing}{tb}{lst}{}
\floatname{listing}{Listing}
\usepackage[dvipsnames,table]{xcolor}
\usepackage{tabularx}
\usepackage{adjustbox}
\usepackage{array}
\usepackage{colortbl}
\usepackage{multirow}
\usepackage{arydshln}
\usepackage{amsthm}
\usepackage{amsmath}

\usepackage{subcaption}
\usepackage{amssymb}
\usepackage{booktabs}

\title{Towards Test-time Efficient Visual Place Recognition \\ via Asymmetric Query Processing}
\author{
    Jaeyoon Kim\equalcontrib, Yoonki Cho\equalcontrib, Sung-Eui Yoon
}
\affiliations{
    Korea Advanced Institute of Science and Technology (KAIST)
}

\begin{document}
\maketitle
\input{sec/0_abstract}    
\input{sec/1_introduction}
\input{sec/2_related_work}
\input{sec/3_method}
\input{sec/4_experiments}
\input{sec/5_conclusion}

\section{Acknowledgements}
This work was supported by the Institute of Information \& communications Technology Planning \& Evaluation~(IITP) grant (No.~RS-2025-25443318, {Physically-grounded Intelligence: A Dual Competency Approach to Embodied AGI through Constructing and Reasoning in the Real World}; No.~RS-2023-00237965, {Recognition, Action and Interaction Algorithms for Open-world Robot Service}) and the National Research Foundation of Korea~(NRF) grant (No.~RS-2023-00208506), both funded by the Korea government~(MSIT).
Prof. Sung-Eui Yoon is a corresponding author (\texttt{e-mail}: \texttt{sungeui@kaist.edu}).
\bibliography{main}

\end{document}

%% file: sec/0_abstract.tex
\begin{abstract}
Visual Place Recognition (VPR) has advanced significantly with high-capacity foundation models like DINOv2, achieving remarkable performance.
Nonetheless, their substantial computational cost makes deployment on resource-constrained devices impractical.
In this paper, we introduce an efficient asymmetric VPR framework that incorporates a high-capacity gallery model for offline feature extraction with a lightweight query network for online processing.
A key challenge in this setting is ensuring compatibility between these heterogeneous networks, which conventional approaches address through computationally expensive k-NN-based compatible training.
To overcome this, we propose a geographical memory bank that structures gallery features using geolocation metadata inherent in VPR databases, eliminating the need for exhaustive k-NN computations. 
Additionally, we introduce an implicit embedding augmentation technique that enhances the query network to model feature variations despite its limited capacity.
Extensive experiments demonstrate that our method not only significantly reduces computational costs but also outperforms existing asymmetric retrieval techniques, establishing a new aspect for VPR in resource-limited environments.
The code is available at \url{https://github.com/jaeyoon1603/AsymVPR}
\end{abstract}

%% file: sec/1_introduction.tex
\begin{figure}[htb!]
    \centering
    \begin{subfigure}[b]{\columnwidth}
        \centering
        \includegraphics[width=0.975\columnwidth]{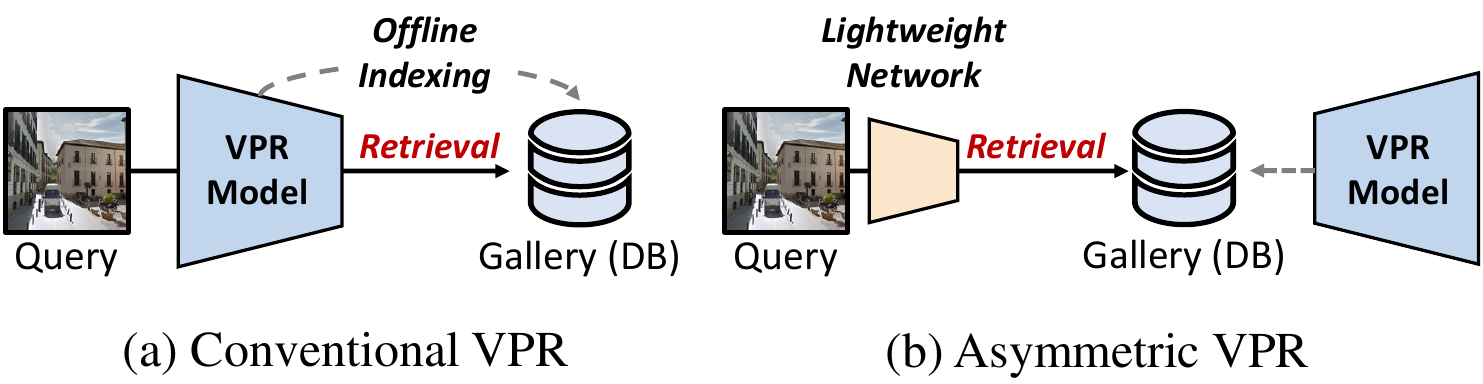}
        \label{fig:fig1-1}
    \end{subfigure}
    \begin{subfigure}[b]{\columnwidth}
        \centering
        \includegraphics[width=0.95\columnwidth]{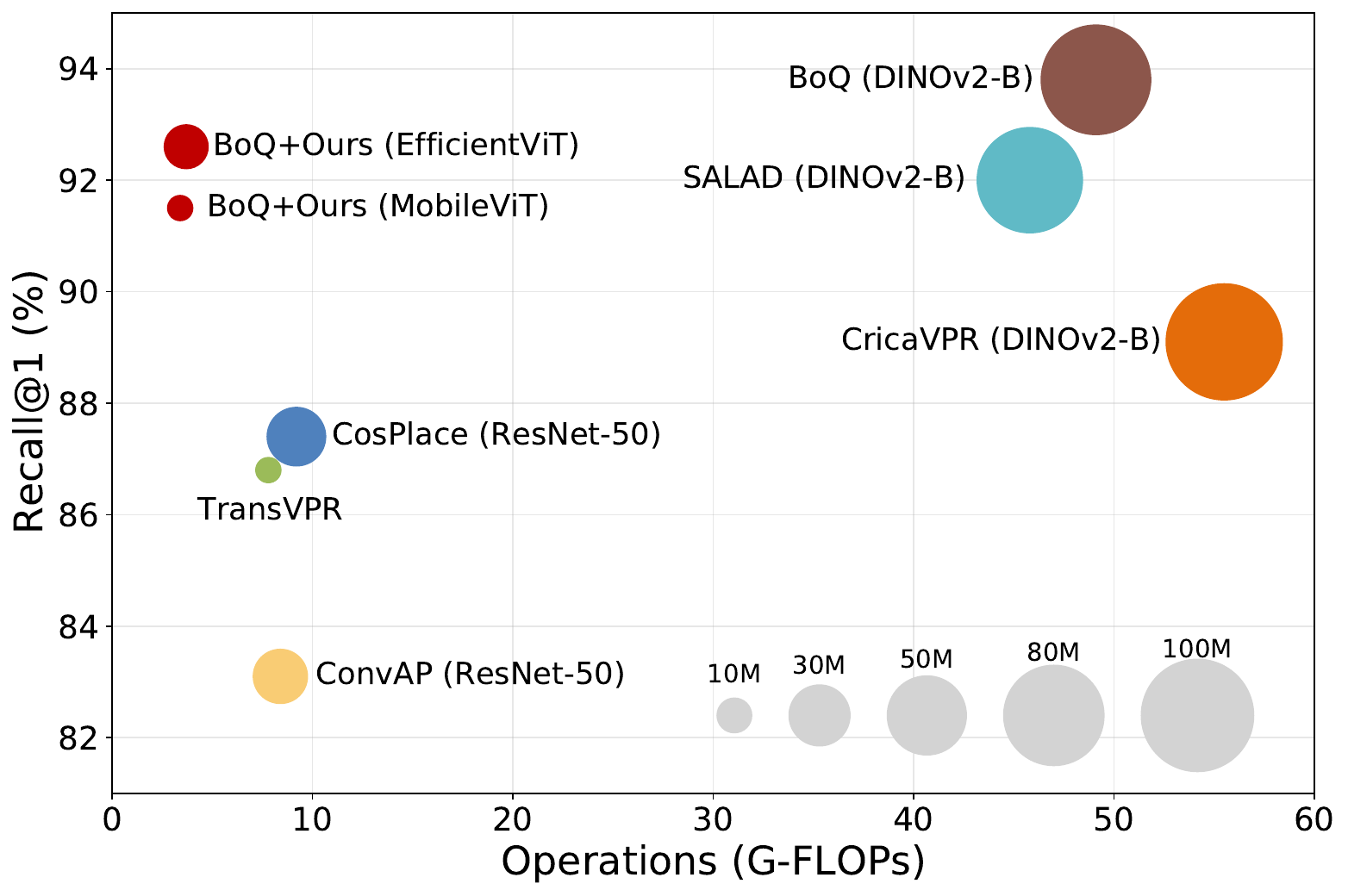}
        \captionsetup{labelformat=empty}
        \caption{(c)}
        \label{fig:fig1-2}
    \end{subfigure}
    \caption{\textbf{Comparison of conventional and asymmetric VPR approaches.}
    (a) Conventional VPR uses the same model for both query and gallery processing. 
    (b) Our asymmetric approach uses a lightweight network for query processing while maintaining a high-capacity model for gallery indexing. 
    (c) Comparison of various VPR methods on Recall@1 vs. computational cost (G-FLOPs) for online query processing on the MSLS validation dataset. Circle sizes indicate the number of model parameters.
    Our method achieves competitive accuracy with significantly less computation, suitable for resource-constrained devices.}
    \label{fig:fig1}
    \vspace{-2mm}
\end{figure}

\section{Introduction}
Visual Place Recognition (VPR) is a fundamental task in computer vision, playing a crucial role in applications such as navigation~\cite{mirowski2018learning,bresson2017simultaneous}, augmented reality~\cite{va_garg2021your}, SLAM~\cite{cadena2016past}, and geo-localization systems~\cite{lowry2015visual, berton2022rethinking}.
It is typically formulated as an image retrieval problem, where a model determines the location of a query image by identifying the most similar geo-tagged reference images in a database (\textit{i.e.,} gallery).
Recent advances in VPR have been driven by foundation models such as DINOv2~\cite{oquab2024dinov}, which have demonstrated remarkable capabilities for rich visual representations~\cite{lu2024towards,lu2024cricavpr, qiu2025emvp, lu2025supervlad}.
Nonetheless, these impressive performance gains come at a substantial computational cost, posing a significant challenge for resource-limited platforms -- deploying such computationally intensive models on resource-constrained devices remains impractical.

To remedy this issue, we explore \textit{asymmetric} retrieval techniques for VPR as a promising and practical solution.
Unlike conventional approaches where the same network processes both query and gallery images, asymmetric retrieval employs separate models with different capacities for each task~\cite{budnik2021asymmetric, duggal2021compatibility}.
While asymmetric retrieval has been explored in other domains~\cite{budnik2021asymmetric, du2024imagesentence, shoshan2024asymmetric, suma2023large}, it remains largely unexplored in VPR, where foundation model-based approaches~\cite{izquierdo2024optimal, tzachor2024effovpr} are now mainstream but operations on edge devices (\textit{e.g.}, robots) are also required.
Fig.~\ref{fig:fig1} illustrates the difference between conventional VPR and our asymmetric framework, showing competitive accuracy with significantly reduced computational requirements.
By leveraging a high-capacity model to precompute and index rich gallery features offline, this approach enables a lightweight query network to efficiently retrieve relevant results during online query processing on resource-constrained environments.

A key challenge in this asymmetric setting is that the query network must produce embeddings that are \textit{compatible} with the gallery’s representation space, despite differences in model capacity and architecture.
Due to the limited capacity of lightweight query models, directly regressing gallery embeddings is insufficient for achieving feature compatibility. 
Recent methods address this limitation by leveraging contextual cues, such as similarity distributions or ranking orders among k-nearest neighbors~(k-NN)~\cite{wu2022contextual, xie2024d3still, wu2023general} relationships.  
However, this approach introduces substantial overhead -- precomputing and storing k-NN information for large-scale gallery embeddings incurs significant computational costs that scale with the size of the database.
In VPR scenarios involving millions of images from diverse locations, these k-NN computations become particularly prohibitive.

In this paper, we introduce an efficient asymmetric framework for VPR that addresses these challenges by leveraging the geospatial structure inherent in visual place recognition.
Unlike prior approaches that rely on k-NN-based contextual information, we propose a \textit{geographical memory bank} that organizes gallery features based on their geographic locations, constructing location-specific statistical representations by aggregating embeddings from images captured within the same vicinity.
For each distinct location, our method computes and stores statistical moments of high-capacity feature distributions, creating a compact yet informative representation of place-specific visual characteristics.
During training, these geographic feature statistics serve as supervisory signals in our asymmetric contrastive learning, guiding the lightweight query network to produce compatible embeddings without requiring exhaustive k-NN searches.
Additionally, we develop an \textit{implicit embedding augmentation} technique that enhances the query network's ability to model feature variations by incorporating location-specific covariance information, enabling effective alignment with rich gallery representations despite limited capacity.
Extensive experiments on standard VPR benchmarks show that our method outperforms state-of-the-art asymmetric retrieval approaches while significantly reducing computational costs for online query processing.

%% file: sec/2_related_work.tex
\section{Related Work}
\label{sec:related_work}
\noindent\textbf{Visual Place Recognition (VPR)} has evolved through significant advancements in feature aggregation techniques.
Early approaches rely on aggregating hand-crafted features (SIFT~\cite{lowe2004distinctive}, SURF~\cite{bay2008speeded}) using methods like Bag of Words~\cite{angeli2008fast} and VLAD~\cite{jegou2010aggregating}.
The emergence of deep learning introduces CNN-based approaches, with NetVLAD~\cite{arandjelovic2016netvlad} pioneering differentiable aggregation for end-to-end training, followed by its variants~\cite{zhang2021vector,yu2019spatial,jin2017learned}. 
Further developments include GeM~\cite{radenovic2018fine} pooling in CosPlace~\cite{berton2022rethinking} and EigenPlaces~\cite{berton2023eigenplaces}, and MLP-based aggregation in MixVPR~\cite{ali2023mixvpr}, all maintaining reasonable computational efficiency.

Recently, foundation models like DINOv2~\cite{oquab2024dinov}, pre-trained with self-supervised learning on diverse visual data, has demonstrated remarkable capabilities with rich visual representations for VPR. 
AnyLoc~\cite{keetha2023anyloc} pioneers the use of DINOv2 features combined with traditional aggregation techniques like VLAD.
Following this paradigm shift, several studies~\cite{lu2024towards, lu2024cricavpr, tzachor2024effovpr,lu2025supervlad, qiu2025emvp} have explored different strategies for employing foundation models' power for VPR. 
SALAD~\cite{izquierdo2024optimal} fine-tunes the pre-trained DINOv2 model specifically for VPR, introducing an optimal transport-based aggregation.
BoQ~\cite{ali2024boq} introduces a novel transformer-based aggregation technique that learns a set of global queries to probe local features via cross-attention.
However, the high computational cost of foundation models poses deployment challenges, as their substantial inference overhead limits practical use on resource-constrained devices.
Our work relaxes these limitations by developing an efficient asymmetric framework that balances performance with computational practicality.

\vspace{0.8mm}
\noindent\textbf{Compatible Training} aims to encode feature embeddings that are interoperable with pre-existing models, facilitating seamless integration across systems for efficient updates~\cite{shen2020towards, wang2020unified, cui2024learning, zhou2022forward}.
This approach significantly reduces the computational burden associated with re-indexing (\textit{i.e.}, backfilling) process when transitioning from older models to newer ones~\cite{su2022privacy, zhang2022hotrefresh, pan2023boundary, bai2022dual, shen2020towards, ramanujan2022forward, meng2021learning}. 
Recent studies~\cite{zhang2022towards, wan2022continual} have explored compatible training in more challenging open-set scenarios, where models adapt to dynamic data distributions while maintaining interoperability.

Building upon this, asymmetric retrieval has emerged as a specialized paradigm within compatible training, deploying a lightweight model on the query side and a more powerful model on the gallery side to balance computational efficiency with retrieval accuracy~\cite{budnik2021asymmetric, duggal2021compatibility, du2024imagesentence}.
\citet{budnik2021asymmetric} first introduced an asymmetric metric learning~(AML) framework and demonstrated that a simple regression loss can distill the embedding spaces.
Recent methods have advanced compatibility between query and gallery models by leveraging k-nearest neighbor (k-NN) information in various ways.
Specifically, CSD~\cite{wu2022contextual} transfers contextual similarity knowledge through k-NN relationships, while D3still~\cite{xie2024d3still} employs decoupled differential distillation based on pairwise similarity across k-NN embeddings.
MSP preserves similarity ranking consistency via monotonic mapping, and its companion approach ROP utilizes smooth Heaviside approximations to constrain sorting results~\cite{wu2023general}. 
In contrast, our method leverages an efficient geographical memory bank with asymmetric contrastive learning, eliminating the need for exhaustive k-NN computations.

%% file: sec/3_method.tex
\begin{figure*}[t]
    \centering
    \includegraphics[width=0.95\textwidth]{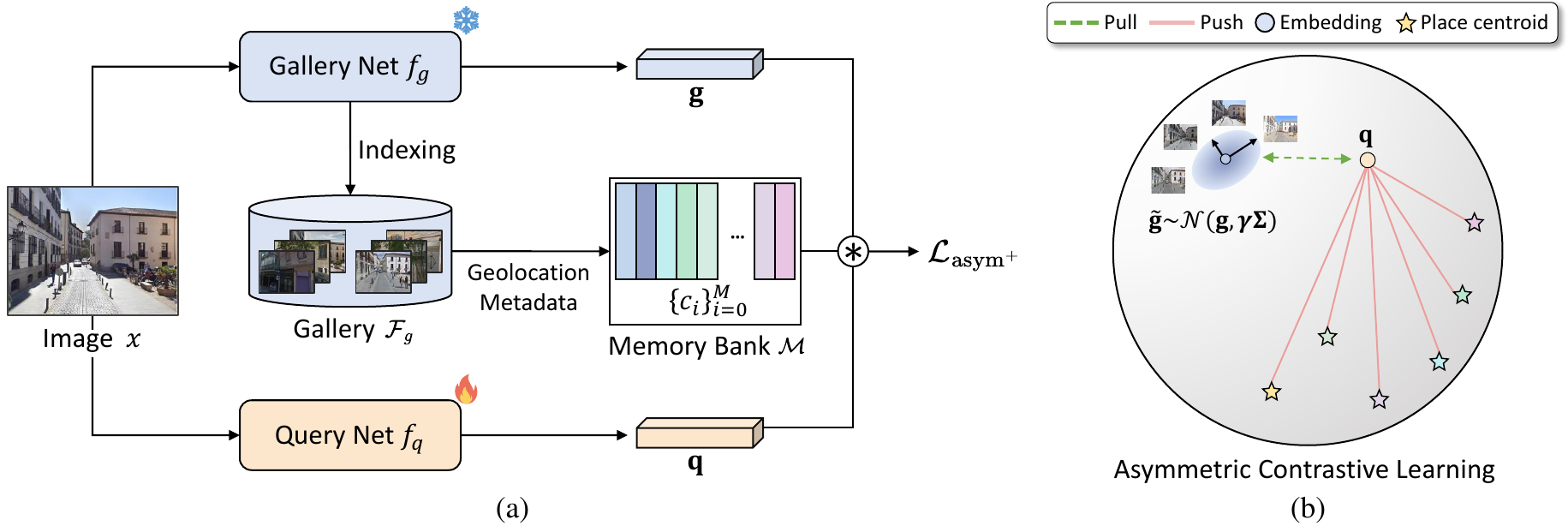}
    \caption{ 
    \textbf{Overview of the proposed asymmetric VPR framework.} 
    (a) A high-capacity gallery network $f_g$ extracts and stores embeddings offline, while a lightweight query network $f_q$ is trained to generate compatible embeddings for efficient online retrieval. The gallery network remains fixed during training while only the query network is optimized.
    The geographical memory bank $\mathcal{M}$ structures gallery features $\mathcal{F}_g$ using geolocation metadata, storing place centroids $\{c_i\}_{i=0}^M$ to efficiently represent feature distributions across locations for compatibility training.
    (b) Illustration of asymmetric contrastive learning with implicit embedding augmentation.
    The query embedding $\mathbf{q}$ is trained to align with its corresponding gallery embedding $\mathbf{g}$ while being pushed away from negative centroids.
    To enhance feature compatibility, augmented gallery embeddings  $\tilde{\mathbf{g}} \sim \mathcal{N}(\mathbf{g}, \gamma\mathbf{\Sigma})$ incorporate location-specific covariance, improving the query network to model feature variations despite its limited capacity.
    }
    \vspace{-2mm}
\label{fig:overview}
\end{figure*}

\section{Method}
\label{sec:method}

The overview of the proposed \textit{asymmetric} VPR framework is illustrated in Fig.~\ref{fig:overview}.
It employs a high-capacity network for gallery feature extraction (performed offline) while utilizing a significantly lighter network for query feature extraction (executed at inference time), enabling efficient retrieval under resource-constrained environments.
In the following subsection, we formally define the problem setting and the constraints imposed by this asymmetric framework.

\subsection{Problem Formulation}
\label{ssec:3-1}

Let $f_g: \mathcal{X} \to \mathbb{R}^d$ be a pre-trained, high-capacity gallery network that has generated a set of features $\mathcal{F}_g = \{\mathbf{g}_i\}_{i=1}^{N}$ extracted from gallery images $\mathcal{X}_g = \{x_i\}_{i=1}^{N}$, where $\mathbf{g}_i = f_g(x_i)$, indexed for retrieval. The query network $f_q: \mathcal{X} \to \mathbb{R}^d$ is a lightweight model designed for efficient inference on resource-constrained devices.

A key challenge in the asymmetric setting—where a lightweight query network must interoperate with a high-capacity gallery network—is that $f_q$ must generate embeddings compatible with $\mathcal{F}_g$, despite differences in model capacity and architecture. 
Training to achieve this compatibility is inherently constrained to using \textit{only the gallery dataset $ \mathcal{X}_g $ and their precomputed features $\mathcal{F}_g$}, without modifying the gallery network or altering the stored embeddings—a practical requirement for real-world deployment where gallery features are precomputed at scale.

Given a query image $x$, we denote its query embedding as $\mathbf{q} = f_q(x)$. Retrieval is performed by ranking the similarity between $\mathbf{q}$ and the gallery embeddings in $\mathcal{F}_g$.
To enable effective retrieval, the following ranking constraint generally needs to be satisfied~\cite{duggal2021compatibility}:
\begin{equation}
    \text{sim}(\mathbf{q}, \mathbf{g}^+) \gg \text{sim}(\mathbf{q}, \mathbf{g}^-),
\end{equation}
where $\mathbf{g}^+$ represents the gallery embedding of a relevant image (corresponding to the same place as the query), $\mathbf{g}^-$ represents the gallery embedding of an irrelevant image (different place), and $\text{sim}(\cdot, \cdot)$ is a similarity function, typically dot product between L2-normalized embeddings.
Satisfying this constraint is challenging since $f_q$ and $f_g$ have different architectures and capacities, potentially leading to disparate embedding spaces.
    
\subsection{Asymmetric Contrastive Learning}
\label{ssec:3-2}
To improve feature compatibility, recent methods have commonly leveraged contextual information from k-nearest neighbor embeddings~\cite{wu2022contextual, wu2023general, xie2024d3still}.  
However, performing k-NN searches and storing this information for all images in a large-scale database incurs a non-negligible computational cost.
To address this, we propose an efficient asymmetric contrastive learning approach that exploits the geographic information naturally available in VPR tasks.
Our approach encourages the query network to produce embeddings that align with those of the gallery network while remaining distinct from features of different places, leveraging global geographic structures rather than local nearest neighbors.

Unlike other retrieval tasks, VPR inherently includes geolocation metadata, which our method leverages without relying on exhaustive nearest neighbor computations.
Specifically, we define our geographical memory bank as a set of $M$ representative centroids:
\begin{equation}
\mathcal{M} = \{\mathbf{c}_j\}_{j=1}^{M} \subset \mathbb{R}^d,
\end{equation}
where each centroid $\mathbf{c}_j$ represents the average of gallery features from $\mathcal{F}_g$ that belong to the same geographic location.
These centroids can be effectively precomputed using GPS coordinates, providing an efficient representation of the feature distribution across different locations.

Given this memory bank, we formulate our asymmetric learning objective as a contrastive learning problem, using geographic centroids as negative samples.
For a given gallery image $x$ with its gallery network embedding $\mathbf{g}$ and query network embedding $\mathbf{q}$, we compute our asymmetric contrastive learning loss by:
\begin{equation}
\label{eq:cl_loss}
\mathcal{L}_{\text{asym}} = -\log \frac{e^{\mathbf{q} \cdot \mathbf{g}/\tau}}
{e^{\mathbf{q} \cdot \mathbf{g}/\tau} + \sum_{j\in\mathcal{N}(x)}e^{\mathbf{q} \cdot \mathbf{c}_j/\tau}},
\end{equation}
where $\tau$ is a temperature parameter that controls the sharpness of the distribution, $\mathcal{N}(x)$ denotes the set of indices for memory bank entries that do not correspond to the same location as image $x$, and $\cdot$ denotes the dot product.

By leveraging geographical information inherent in VPR datasets, our approach eliminates the need for exhaustive k-NN computations, significantly reducing computational overhead while scaling efficiently with database size.
Additionally, our geographically informed memory bank captures global feature relationships, leveraging priors tailored for VPR to enhance feature compatibility.

\subsection{Implicit Embedding Augmentation}
\label{ssec:3-3}
While our asymmetric contrastive learning framework considers geographical feature relationships, it can still face challenges caused by the capacity gap between the query network and the gallery network. 
Specifically, the lightweight query model may struggle to fully capture the complex distribution and variability of features within individual locations that the high-capacity gallery model can encode.
To address this, we introduce an implicit embedding augmentation technique that exploits location-specific feature variations, enhancing the compatibility to approximate the rich representations of the gallery network.

Embedding augmentation is typically modeled using multivariate normal distribution~\cite{han2023clothing,wang2019implicit,li2021simple}, which we adopt in our formulation:
\begin{equation}
\tilde{\mathbf{g}} \sim \mathcal{N}(\mathbf{g}, \gamma\Sigma),
\end{equation}
where $\Sigma \in \mathbb{R}^{d \times d}$ represents the covariance matrix of features in location of $\mathbf{g}$, and $\gamma$ is a scaling hyperparameter.

To incorporate this augmentation into our training framework, we modify our contrastive learning loss (Eq.~\ref{eq:cl_loss}) by replacing the single gallery embedding with multiple augmented samples. 
For each gallery image, we can generate K augmented embeddings $\{\tilde{\mathbf{g}}^1, \tilde{\mathbf{g}}^2, ..., \tilde{\mathbf{g}}^K\}$ from the defined multivariate normal distribution. We then compute the explicit augmentation loss for the augmented samples:
\begin{equation}
\label{eq:explicit}
\mathcal{L}^{\text{exp}}_{\text{asym}^+} = -\frac{1}{K} \sum_{k=1}^{K} \log 
\frac{e^{\mathbf{q} \cdot \tilde{\mathbf{g}}^k / \tau}}
{e^{\mathbf{q} \cdot \tilde{\mathbf{g}}^k / \tau} + \sum_{j\in\mathcal{N}(x)} e^{\mathbf{q} \cdot \mathbf{c}_j / \tau}}.
\end{equation}
\noindent\textbf{Proposition 1.} \textit{The explicit augmentation loss can be transformed into an implicit form as $K \rightarrow \infty$.}
\begin{proof}
When $K \rightarrow \infty$, Eq.~\ref{eq:explicit} becomes an expectation:
\begin{equation}
\bar{\mathcal{L}}_{\text{asym}^+} = \mathbb{E}_{\tilde{\mathbf{g}}}\left[\log\left(1 + \sum_{j\in\mathcal{N}(x)} e^{\mathbf{q} \cdot \mathbf{c}_j / \tau - \mathbf{q} \cdot \tilde{\mathbf{g}} / \tau}\right)\right].
\end{equation}
Since the logarithm function is concave, we can apply Jensen's inequality:
\begin{equation}
\label{eq7}
\begin{aligned}
\bar{\mathcal{L}}_{\text{asym}^+} &\leq \log\left(1 + \sum_{j\in\mathcal{N}(x)} \mathbb{E}_{\tilde{\mathbf{g}}}\left[e^{\mathbf{q} \cdot \mathbf{c}_j / \tau - \mathbf{q} \cdot \tilde{\mathbf{g}} / \tau}\right]\right)\\
&=\log\left(1 + \sum_{j\in\mathcal{N}(x)} e^{\mathbf{q} \cdot \mathbf{c}_j / \tau - \mathbf{q} \cdot \mathbf{g} / \tau + \frac{\gamma}{2\tau^2} \mathbf{q}^T \Sigma \mathbf{q}}\right),
\end{aligned}
\end{equation}
where moment-generating function of multivariate Gaussian is used as:
$\mathbb{E}[e^{-\mathbf{q} \cdot \tilde{\mathbf{g}} / \tau}] = e^{-\mathbf{q} \cdot \mathbf{g} / \tau + \frac{\gamma}{2\tau^2} \mathbf{q}^T \Sigma \mathbf{q}}$.
\end{proof}
For simplicity, our final implicit embedding augmentation loss can be defined by using the upper bound in Eq.~\ref{eq7}:
\begin{equation}
\label{eq:our_loss}
\mathcal{L}_{\text{asym}^+} = -\log \frac{e^{\mathbf{q} \cdot \mathbf{g}/\tau}}
{e^{\mathbf{q} \cdot \mathbf{g}/\tau} + \sum_{j\in\mathcal{N}(x)}e^{\mathbf{q} \cdot \mathbf{c}_j/\tau + (\gamma/2\tau^2)\mathbf{q}^T \Sigma \mathbf{q}}}.
\end{equation}
The formulation introduces a variance-based regularization term $(\gamma/2\tau^2)\mathbf{q}^T \Sigma \mathbf{q}$ for the negative terms, incorporating covariance information without requiring explicit sampling.

\vspace{1mm}
\noindent\textbf{Regularization effect of covariance guidance.}\hspace{1mm}
Using the eigendecomposition, the quadratic form becomes the following.
\begin{equation}
\begin{aligned}
\mathbf{q}^T \Sigma \mathbf{q} &= \mathbf{q}^T V\Lambda V^T \mathbf{q} \\
&= (V^T \mathbf{q})^T \Lambda (V^T \mathbf{q}) \\
&= \sum_j \lambda_j (v_j^T \mathbf{q})^2,
 \end{aligned}
\end{equation}
where $\lambda_j$ and $v_j$ are the $j$-th eigenvalue and eigenvector, respectively.
To minimize Eq.~\ref{eq:our_loss}, the regularization term must also be minimized during the optimization process, since it is attached to the negative terms. 
It results in an optimization behavior where the query model takes stronger penalties along directions with larger eigenvalues (\textit{i.e.}, higher variance) in the gallery's feature space.

These high-variance directions capture less discriminative intra-place variations, such as changing viewpoints or lighting. By reducing sensitivity to these, our approach guides the lightweight query model to prioritize stable, place-distinctive features, enabling efficient asymmetric retrieval despite its limited capacity.

%% file: sec/4_experiments.tex
\section{Experiments}
\subsection{Experimental Setup}
\label{ssec:4-1}
\noindent\textbf{Training and Evaluation Protocols.}\hspace{1mm}
For training, we utilize GSV-Cities~\cite{ali2022gsv} dataset, a large-scale collection of 560k images from 67k distinct locations spanning a 14-year period, sourced from Google Street View. 
To construct the gallery embeddings, we preprocess the entire database by extracting features from all images using a gallery network. These precomputed gallery features are stored offline and remain fixed throughout training, aligning with the practical constraints of real-world deployment. We then utilize the geolocation metadata (\textit{i.e.}, GPS coordinates) to compute place centroids by averaging the gallery embeddings from the same location. Simultaneously, we compute the covariance matrices for each place, which are used for implicit embedding augmentation.

For evaluation, we use five standard VPR benchmarks: Pitts250k~\cite{torii2013visual}, MSLS Validation~\cite{warburg2020mapillary},  Tokyo24/7~\cite{torii201524}, Nordland~\cite{zaffar2021vpr}, and, AmsterTime~\cite{yildiz2022amstertime} which cover diverse challenges such as viewpoint shifts, seasonal variations, and illumination changes. Following prior VPR literature~\cite{lu2024towards, lu2024cricavpr,izquierdo2024optimal}, we evaluate performance using the Recall@k (R@k) as an evaluation metric, which measures the percentage of queries with at least one correct match among the top-K retrieved candidates. 
Positive matches are determined using a 25-meter threshold based on geographic coordinates, except for Nordland, where a match is considered correct if retrieved within three frames of the ground truth position, and AmsterTime which utilizes its provided ground truth pairs.

\vspace{1.0mm}
\noindent\textbf{Network Architectures.}\hspace{1mm}
For the gallery models, we use SALAD~\cite{izquierdo2024optimal} and BoQ~\cite{ali2024boq}, both trained on GSV-Cities and built on DINOv2-B, providing strong visual representations with embedding dimensions of 8448 and 12288, respectively. 
These high-capacity models impose substantial computational demands, making them a practical choice for our asymmetric setting, where gallery features are precomputed offline.
For the query models, we use two popular lightweight backbones: MobileViTv2~\cite{mehtaseparable} and EfficientViT-B2~\cite{cai2023efficientvit}, designed for mobile and resource-constrained devices. 
To extract features with these lightweight backbones, we maintain architectural consistency by employing the same type of aggregator on both gallery and query sides in a symmetric manner. For instance, when SALAD serves as the gallery model, its aggregation architecture is also used in the query model.

\vspace{1.0mm}
\noindent\textbf{Implementation Details.}\hspace{1mm}
We train our models using the AdamW~\cite{loshchilov2017decoupled} optimizer with an initial learning rate of $5\times10^{-4}$ and cosine learning rate decay, reaching a minimum learning rate of $1\times10^{-4}$. Training is performed for 15 epochs with a batch size of 64.
All images are resized to 322×322 for both training and evaluation. 
We empirically set the temperature parameter $\tau = 0.05$ for asymmetric contrastive learning, and the scaling parameter $\gamma = 15$ for implicit embedding augmentation.
Utilizing the full covariance matrix during training incurs GPU memory overhead due to the high dimensionality of the feature space. To mitigate this, we approximate it using only the diagonal elements of the covariance matrix, significantly reducing computational costs while preserving essential statistical information.
We implement our framework in PyTorch~\cite{paszke2019pytorch} and utilize a single RTX-4090 GPU for all experiments.

\begingroup
\renewcommand{\arraystretch}{1.10}  
\begin{table*}[t]
\centering
\begin{adjustbox}{max width=\textwidth}
\begin{tabular}{lc*{10}{>{\centering\arraybackslash}p{0.9cm}}}
\toprule
\multirow{2}{*}{Method} & \multirow{2}{*}{Query Network} &\multicolumn{2}{c}{Pitts250k}&\multicolumn{2}{c}{MSLS Val}&\multicolumn{2}{c}{Tokyo24/7}&\multicolumn{2}{c}{Nordland}&\multicolumn{2}{c}{AmsterTime}\\
& & R@1&R@5& R@1&R@5& R@1&R@5& R@1&R@5& R@1&R@5\\
\hline
\rowcolor{gray!10}\multicolumn{12}{c}{\textit{Training with \textbf{SALAD} of DINOv2 as gallery model}}\\
SALAD\textsuperscript{$\dagger$}~\cite{izquierdo2024optimal} & DINOv2-B & 95.1&98.6& 92.0&96.4& 94.6 & 97.5 & 76.0 & 89.2 & 58.5 & 78.9 \\
MobileViTv2-SALAD\textsuperscript{$\dagger$} &MobileViTv2& 87.9 & 95.1 & 76.6 & 85.0 & 70.5 & 83.5 & 18.3 & 28.8 & 28.5 & 45.6 \\
EfficientViT-B2-SALAD\textsuperscript{$\dagger$} &EfficientViT-B2& 92.1 & 97.1 & 85.3 & 91.4 & 83.2 & 92.1 & 26.5 & 39.1 & 38.5 & 59.5 \\
\hdashline
CSD~\cite{wu2022contextual}&\multirow{5}{*}{MobileViTv2}&88.1&95.5&87.3&94.7&75.2&87.3&38.8&56.2&34.9 &54.8 \\
ROP~\cite{wu2023general}&& 86.5&94.5&87.3&94.3&69.2&81.6&32.9&50.4&34.8 &56.0 \\
MSP~\cite{wu2023general}&&88.4&\textbf{95.6}&88.5&94.5&76.2&\textbf{87.9}&41.0&57.0&34.8 &55.5 \\
D3still~\cite{xie2024d3still}&& 86.5& 94.0& 87.8& 94.6& 71.4& 84.1& 41.3& 57.5&33.3  &53.0 \\
\textbf{Ours} &&\textbf{89.0} & 95.2 & \textbf{89.3} & \textbf{95.0} & \textbf{76.5} & 87.6 & \textbf{49.3} & \textbf{68.0} & \textbf{37.6} & \textbf{56.1}\\
\hdashline
CSD~\cite{wu2022contextual}&\multirow{5}{*}{EfficientViT-B2}&91.2&97.0&88.8&95.0&82.2&91.7&53.6&71.6&37.4 &57.3 \\
ROP~\cite{wu2023general}&& 89.1& 96.1& 87.8& 94.6& 72.1& 86.0& 44.5& 61.7&35.1  &55.6 \\
MSP~\cite{wu2023general}&&91.6&96.7&88.2&95.4&85.7& 94.2&55.5&73.2&41.8 &62.2 \\
D3still~\cite{xie2024d3still}&& 91.9 & 97.0 & 89.7 & 95.7 & 85.4 & \textbf{94.3} & 58.2 & 75.8 &40.9  &61.3 \\
\textbf{Ours}&& \textbf{92.4} & \textbf{97.5} & \textbf{91.2} & \textbf{95.8} & \textbf{86.7} & 94.0 & \textbf{61.7} & \textbf{77.6} & \textbf{43.0} & \textbf{64.5} \\
\hline
\rowcolor{gray!10}\multicolumn{12}{c}{\textit{Training with \textbf{BoQ} of DINOv2 as gallery model }}\\
BoQ\textsuperscript{$\dagger$}~\cite{ali2024boq} & DINOv2-B &96.6 &99.1 &93.8 &96.8 & 96.5&97.8&81.3&92.5 & 63.0 & 81.6 \\
MobileViTv2-BoQ\textsuperscript{$\dagger$} & MobileViTv2 & 92.8 & 98.0 & 85.0 & 92.3 & 85.1 & 93.3 & 49.4 & 65.5 & 38.7 & 58.0\\
EfficientViT-B2-BoQ\textsuperscript{$\dagger$} & EfficientViT-B2 & 94.3 & 98.3 & 87.7 & 93.2 & 85.7 & 92.7 & 51.6 & 67.3 & 41.6 & 63.8 \\
\hdashline
CSD~\cite{wu2022contextual}& \multirow{5}{*}{MobileViTv2} &92.5&97.5&90.9&95.7&85.7&93.7&62.6&79.5&42.6 &64.2 \\
ROP~\cite{wu2023general}&& 90.7& 96.8& 89.6& 95.4& 81.3& 89.5& 56.9& 75.8& 39.6 &62.9 \\
MSP~\cite{wu2023general}&&93.0&97.6&90.0&95.4&87.0&93.3&64.9&81.1& 41.7&63.0 \\
D3still~\cite{xie2024d3still}&& 93.2 & 97.5 & 90.4 & 95.5 & 87.0 & 93.7 & 66.5 & 82.2 &41.7  &64.3 \\
\textbf{Ours} && \textbf{93.4} & \textbf{97.8} & \textbf{91.1} & \textbf{96.1} & \textbf{89.2} & \textbf{94.3} & \textbf{67.8} & \textbf{82.8} &\textbf{43.8} & \textbf{67.0} \\
\hdashline
CSD~\cite{wu2022contextual}& \multirow{5}{*}{EfficientViT-B2} &94.6&98.4&91.4&\textbf{96.5}&90.5&94.9&68.7&82.4&48.1 &69.6 \\
ROP~\cite{wu2023general}&& 91.6 & 97.1& 90.5 & 96.0 & 84.1& 91.4& 60.3& 77.1&  37.4&58.6 \\
MSP~\cite{wu2023general}&& 94.3 & 98.1 & 91.4 & 95.8 & 90.5 & 94.3 & 69.0 & 85.0 &  46.5&67.9 \\
D3still~\cite{xie2024d3still}&& 95.0& 98.4& 91.9& 95.9& 91.7& 95.9& 72.8& 86.9&46.7  &69.9 \\
\textbf{Ours} && \textbf{95.4} & \textbf{98.5} & \textbf{92.3} & 95.9 & \textbf{92.7} & \textbf{96.5} & \textbf{74.6} & \textbf{88.3} & \textbf{48.6} & \textbf{71.3} \\
\bottomrule
\end{tabular}
\end{adjustbox}
\vspace{-0.5mm}\caption{Comparison with state-of-the-art asymmetric retrieval methods on standard VPR benchmarks. $\dagger$ denotes performance using the same model for both query and gallery embeddings (\textit{i.e.,} symmetric retrieval).}\vspace{-2mm}
\label{tab:sota}
\end{table*}
\endgroup

\subsection{Comparison with State-of-the-Arts}

We evaluate our method against state-of-the-art asymmetric retrieval approaches, including CSD~\cite{wu2022contextual}, ROP~\cite{wu2023general}, MSP~\cite{wu2023general}, and D3still~\cite{xie2024d3still}, across standard VPR benchmarks. The results are summarized in Table~\ref{tab:sota}. The table is structured into two primary sections, each dedicated to one of the aggregators: SALAD~\cite{izquierdo2024optimal} and BoQ~\cite{ali2024boq}. In each section, the upper part details the symmetric retrieval performance. This includes the result of the high-capacity model itself (e.g., BoQ with DINOv2-B), which can be regarded as the performance upper bound. Additionally, it provides baseline results for lightweight models (MobileViTv2~\cite{mehtaseparable} and EfficientViT-B2~\cite{cai2023efficientvit}) when used symmetrically for both query and gallery processing. The remainder of each section presents the core asymmetric retrieval results. In this configuration, the high-capacity gallery model is fixed, while different lightweight query networks are trained for compatibility.

As shown in the table, our method achieve superior performance~(R@1) against existing asymmetric retrieval approaches across all datasets and query-gallery model configurations. Specifically, when using BoQ as the gallery model and EfficientViT-B2 as the query network, it achieves strong retrieval performance, reaching 95.4\% R@1 on Pitts250k, 92.3\% on MSLS Val, and 92.7\% on Tokyo24/7. Additionally, our method demonstrates consistent improvements on Nordland and AmsterTime, achieving 74.6\% and 48.6\%, respectively, both of which present diverse environmental variations inherently challenging for VPR. These results highlight the effectiveness of leveraging the geographical prior knowledge inherent in VPR, bridging the capacity gap and improving compatibility between lightweight query networks and high-capacity gallery models. By incorporating this structure through the geographical memory bank with implicit embedding augmentation, our method enhances feature compatibility while mitigating the computational overhead of k-NN-based techniques.

\subsection{Analysis and Ablation Study}
We conduct extensive analysis and ablation studies to validate the effectiveness of our method and design choices. Unless otherwise specified, all experiments use BoQ as a gallery model and EfficientViT-B2 as a query model.

\begingroup
\renewcommand{\arraystretch}{1.15}  
\begin{table}[t]
    \centering
    \begin{adjustbox}{width=1\linewidth}
    \begin{tabular}{ccrcrccr}
        \toprule
        & \multirow{2}{*}{Backbone} & \multicolumn{2}{c}{FLOPs} & \multicolumn{2}{c}{Params} & \multicolumn{2}{c}{Latency} \\
        \cmidrule(lr){3-4} \cmidrule(lr){5-6} \cmidrule(lr){7-8}
        & & (G) & rel. & (M) & rel. & (ms) & rel. \\
        \midrule
        \multirow{3}{*}{\rotatebox[origin=c]{90}{SALAD}} 
        & DINOv2-B & 45.8 & 100\% & 88.0 & 100\% & 61.5& 1.0× \\
        & EfficientViT-B2 & 3.7& 8.1\% &15.8 & 18.0\% & 17.8& 3.5× \\
        & MobileViTv2 & 3.4& 7.4\% & 5.4& 6.1\% & 15.2& 4.0× \\
        \midrule
        \multirow{3}{*}{\rotatebox[origin=c]{90}{BoQ}} 
        & DINOv2-B & 49.1& 100\% & 95.2& 100\% & 62.4& 1.0× \\
        & EfficientViT-B2 & 4.4& 8.9\% & 22.3& 23.4\% & 21.1& 3.0× \\
        & MobileViTv2 & 4.2& 8.5\% & 12.1& 12.7\% & 17.3& 3.6× \\
        \bottomrule
    \end{tabular}
    \end{adjustbox}
    \vspace{-0.5mm}\caption{Computational efficiency comparison of high-capacity and lightweight backbones.}\vspace{-2mm}
    \label{tab:online_cost}
\end{table}
\endgroup

\vspace{1.0mm}
\noindent\textbf{Computational Efficiency Analysis.}\hspace{1mm}
We analyze the computational efficiency for online query processing of different backbone and aggregator configurations in our asymmetric VPR framework, with the results summarized in Table~\ref{tab:online_cost}. Compared to high-capacity gallery networks, lightweight query networks offer substantial efficiency improvements, highlighting the effectiveness of asymmetric VPR framework. For instance, EfficientViT-B2 requires only 7.3\% of the FLOPs and 18.0\% of the parameters compared to DINOv2-B when used with SALAD, while enabling 3.5× faster embedding extraction time. MobileViTv2 achieves even greater efficiency, requiring only 6.7\% of FLOPs and 6.1\% of parameters, leading to 4.0× faster inference. With similar efficiency gains with BoQ, these significant reductions in computational demands make real-time VPR feasible on mobile and edge devices with limited capabilities.

\begingroup
\renewcommand{\arraystretch}{1.10}  
\begin{table}[t]
    \centering
        \begin{adjustbox}{width=\columnwidth}
        \begin{tabular}{*{1}{p{1.2cm}} *{3}{>{\centering\arraybackslash}p{2.5cm}}} 
            \toprule
            \multirow{2}{*}{Method} & Precomputation & Training Iteration & GPU Memory \\
            & (min) & (sec) & (GB) \\
            \midrule
            CSD & 1392.76 & 1.34 & 23.9 \\
            Ours & 0.26   & 0.19 & 17.5 \\
            \bottomrule
        \end{tabular}
    \end{adjustbox}
    \vspace{-0.5mm}\caption{Computational efficiency comparison of asymmetric training between CSD~\cite{wu2022contextual} and our method.}\vspace{-2mm}
    \label{tab:offline_cost}
\end{table}

\begingroup
\renewcommand{\arraystretch}{1.10}  
\begin{table*}[t]
\centering
\begin{adjustbox}{max width=\textwidth}
\begin{tabular}{lc*{10}{>{\centering\arraybackslash}p{1.1cm}}}
\toprule
\multirow{2}{*}{Method} &\multicolumn{2}{c}{Pitts250k}&\multicolumn{2}{c}{MSLS Val}&\multicolumn{2}{c}{Tokyo24/7}&\multicolumn{2}{c}{Nordland}&\multicolumn{2}{c}{AmsterTime}\\
&R@1&R@5& R@1&R@5& R@1&R@5& R@1&R@5& R@1&R@5\\
\midrule
Ours&\textbf{95.4} & \textbf{98.5} & \textbf{92.3} & \textbf{95.9} & \textbf{92.7} & \textbf{96.5} & \textbf{74.6} & \textbf{88.3} & \textbf{48.6} & \textbf{71.3} \\
\midrule
\textit{without} implicit embedding augmentation & 94.3 & 98.1 & 91.2 & 95.7 &  90.2 & 95.6 & 70.3 & 85.3 & 46.9 & 68.9 \\ 
\textit{with} explicit embedding augmentation &94.7&98.4&91.5&96.2&91.4&95.2&73.3&86.8& 47.7 & 70.8 \\
\textit{with} queue-based memory bank &94.2 & 97.9 & 91.4 & 96.1 & 90.2 & 95.9 & 72.0 & 85.8 & 47.1 & 69.7 \\
\bottomrule
\end{tabular}
\end{adjustbox}
\vspace{-0.5mm}\caption{Ablation study with BoQ as a gallery model and EfficientViT-B2 as a query model, comparing our method to variants without implicit embedding augmentation, with explicit embedding augmentation, and with a queue-based memory bank.}\vspace{-2mm}
\label{tab:ablation_study}
\end{table*}
\endgroup

Additionally, we analyze the asymmetric training efficiency of our method compared to CSD~\cite{wu2022contextual}, a representative k-NN-based method, in Table~\ref{tab:offline_cost}.  Our approach reduces the gallery embedding precomputation time from 1392.76 minutes to just 0.26 minutes. This significant reduction is achieved by replacing exhaustive k-NN computations with a geographical memory bank leveraging the geolocation metadata inherent in VPR datasets. Training iterations are also faster, as our method removes k-NN processing overhead and lowers GPU memory consumption. These improvements are particularly significant for large-scale VPR applications, where the computational cost of k-NN based methods grows with database size, while our geolocation-based approach maintains consistent efficiency.

\vspace{1mm}
\noindent\textbf{Ablation Study.}\hspace{1mm}
To evaluate the effectiveness of each component in our framework, we conduct an ablation study, with the results summarized in Table~\ref{tab:ablation_study}. Specifically, removing implicit embedding augmentation~(\textit{i.e.}, applying $\mathcal{L}_{\text{asym}}$ in Eq.~\ref{eq:cl_loss}) leads to a consistent performance drop across all benchmarks, with R@1 decreasing by up to 4.3\% on the challenging Nordland dataset, demonstrating its role in bridging the capacity gap between query and gallery networks. We further compare our implicit augmentation with explicit embedding augmentation in Eq.~\ref{eq:explicit}, where $K$ is set to 10. While explicit augmentation provides some improvement over the baseline, it remains less effective than the implicit formulation, highlighting the advantage of leveraging feature statistics. Additionally, replacing our geographical memory bank with a queue-based memory bank, which maintains a fixed-size queue of past embeddings as a dynamic dictionary~\cite{he2020momentum}, results in overall performance degradation, with the drop of 2.6\% R@1 on Nordland. This highlights the advantage of structured geospatial priors, which are particularly crucial for VPR, over dynamically updated feature queues. Overall, these results validate the effectiveness of each component in our proposed framework for asymmetric VPR performance.

\begin{table}[t!]
    \centering
    \begin{adjustbox}{max width=\columnwidth}
    \begin{tabular}{*{1}{>{\centering\arraybackslash}p{2.3cm}} p{1.45cm} *{3}{>{\centering\arraybackslash}p{1.55cm}}} 
            \toprule
            Aggregator & Method & Tokyo24/7 & Nordland & AmsterTime \\
            \midrule
            \multirow{4}{*}{\shortstack{MLP\vspace{0.5mm}\\ (33.4M)}} 
            & CSD & 81.6 & 39.3 & 36.1 \\
            & MSP & 81.0& 40.1 & 34.8 \\
            & D3still & 86.9 & 37.2 & 35.5 \\
            & \textbf{Ours} & 89.1 & 43.6 & 38.7 \\
            \midrule
            \multirow{4}{*}{\shortstack{Symmetric (Ours)\vspace{0.5mm}\\ (22.3M)}} 
            & CSD & 90.5 & 68.7 & 48.1 \\
            & MSP & 90.5& 69.0 & 46.5 \\
            & D3still & 91.7 & 72.8 & 46.7 \\
            & \textbf{Ours} & 92.7 & 74.6 & 48.6 \\
            \bottomrule
        \end{tabular}
    \end{adjustbox}
    \vspace{-0.5mm}\caption{Ablation study  (R@1\%) of aggregation strategies between MLP-based and Symmetric (Ours) architectures.}\vspace{-2mm}
    \label{tab:aggregator}
\end{table}
\endgroup

\vspace{1.0mm}
\noindent\textbf{Impact of Aggregation Architecture.}\hspace{1mm}
We apply the same type of aggregator as the gallery network to extract features from the lightweight query network, ensuring better compatibility with the gallery model. To analyze the impact of maintaining architectural consistency of aggregators, we compare our symmetric aggregation approach with an alternative setup that employs an MLP-based transformation to align with the gallery embedding space. Despite having significantly more parameters (33.4M vs. 22.3M), the MLP-based approach leads to lower asymmetric retrieval performance, as shown in Table~\ref{tab:aggregator}. 
This finding underscores the importance of structural alignment in feature aggregation for asymmetric VPR, suggesting that architectural consistency between query and gallery models plays a crucial role in improving feature compatibility.

\begin{figure}[t]
    \centering
    \includegraphics[width=0.475\textwidth]{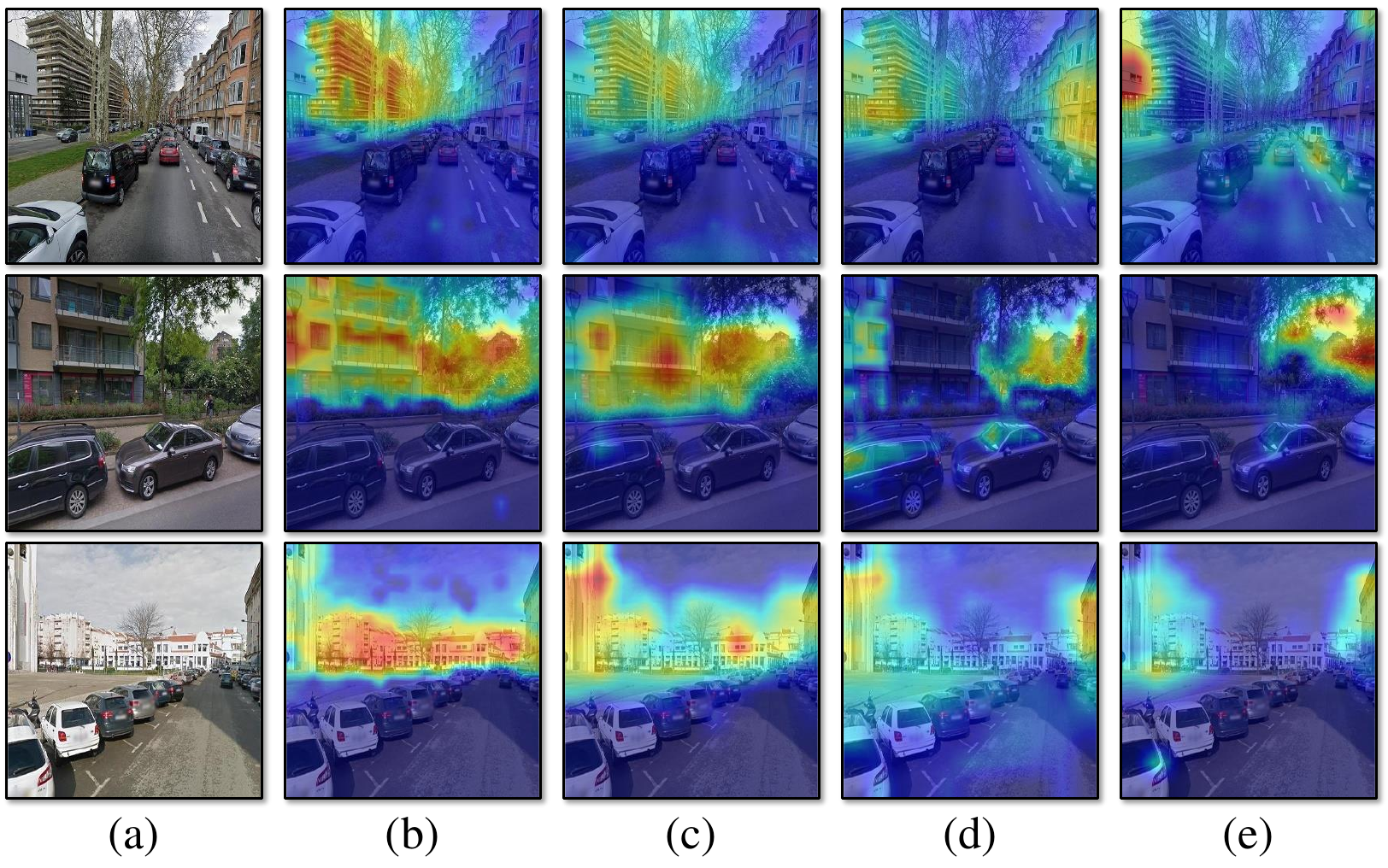}
    \caption{Grad-CAM~\cite{selvaraju2017grad} visualization using geographical memory bank as a classifier.
    (a) Original image; (b) Gallery model (\textit{i.e.,} BoQ); (c) Query model by ours; (d) Query model by ours without implicit embedding augmentation; (e) Query model by CSD~\cite{wu2022contextual}.}\vspace{-2mm}
    \label{fig:gradcam}
\end{figure}

\vspace{1.0mm}
\noindent\textbf{Qualitative Analysis.}\hspace{1mm}
To provide deeper insights into the effectiveness of our method, we conduct qualitative analysis using Grad-CAM~\cite{selvaraju2017grad} with the geographical memory bank as a linear classifier. 
As shown in Fig.~\ref{fig:gradcam}, different methods exhibit distinct attention patterns, revealing key insights about feature alignment.
The high-capacity gallery model (Fig.3(b)) demonstrates rich attention patterns capturing fine-grained scene details. Our method (Fig.3(c)) successfully emulates these patterns, achieving strong alignment with the gallery model's focus regions while maintaining efficiency. In contrast, our variant without implicit embedding augmentation (Fig.3(d)) exhibits scattered attention patterns, highlighting the critical role of our augmentation technique in bridging the capacity gap. The k-NN-based CSD (Fig.3(e)) shows markedly different attention patterns with less structured emphasis regions, suggesting insufficient feature compatibility.
This analysis supports our quantitative findings, showing that our method achieves better feature compatibility than existing asymmetric approaches.

%% file: sec/5_conclusion.tex
\section{Conclusion}
In this paper, we introduced an efficient asymmetric framework for visual place recognition (VPR), where a lightweight query network is trained to be compatible with a high-capacity gallery model, enabling efficient retrieval in resource-constrained environments. We proposed a geographical memory bank that eliminates the need for exhaustive k-NN computations, and an implicit embedding augmentation technique that bridges the capacity gap between query and gallery models. Extensive experimental results demonstrated that our method not only significantly reduces computational costs compared to conventional VPR methods but also consistently outperforms existing approaches.